\def\year{2021}\relax
\documentclass[letterpaper]{article} 
\usepackage{aaai21}  
\usepackage{times}  
\usepackage{helvet} 
\usepackage{courier}  
\usepackage[hyphens]{url}  
\usepackage{graphicx} 
\usepackage{natbib}  
\usepackage{caption} 
\usepackage{times}
\usepackage{latexsym}
\usepackage{url}
\usepackage{amsmath}
\usepackage{amssymb}
\usepackage{multirow}
\usepackage{graphicx}
\usepackage{booktabs}

\usepackage{color}
\usepackage[switch]{lineno}

\title{Document-level Neural Machine Translation with Document Embeddings}
\author{
    Shu Jiang\textsuperscript{\rm 1,2,3}, 
    Hai Zhao\textsuperscript{\rm 1,2,3}
    \thanks{Corresponding author. This paper was partially supported by National Key Research and Development Program of China (No. 2017YFB0304100), Key Projects of National Natural Science Foundation of China (U1836222 and 61733011), Huawei-SJTU long term AI project, Cutting-edge Machine reading comprehension and language  model.},
    Zuchao Li\textsuperscript{\rm 1,2,3}, 
    Bao-Liang Lu\textsuperscript{\rm 1,2,3}
    \\
}
\affiliations{
    \textsuperscript{\rm 1} Department of Computer Science and Engineering, Shanghai Jiao Tong University \\
    \textsuperscript{\rm 2} Key Laboratory of Shanghai Education Commission for Intelligent Interaction \\
    and Cognitive Engineering, Shanghai Jiao Tong University, Shanghai, China \\
    \textsuperscript{\rm 3} MoE Key Lab of Artificial Intelligence, AI Institute, Shanghai Jiao Tong University, Shanghai, China

    jshmjs45@gmail.com, zhaohai@cs.sjtu.edu.cn, charlee@sjtu.edu.cn, bllu@sjtu.edu.cn

}

\begin{document}

\maketitle

\begin{abstract}
Standard neural machine translation (NMT) is on the assumption of document-level context independent. Most existing document-level NMT methods are satisfied with a smattering sense of brief document-level information, while this work focuses on exploiting detailed document-level context in terms of multiple forms of document embeddings, which is capable of sufficiently modeling deeper and richer document-level context. The proposed document-aware NMT is implemented to enhance the Transformer baseline by introducing both global and local document-level clues on the source end. Experiments show that the proposed method significantly improves the translation performance over strong baselines and other related studies.
\end{abstract}

\section{Introduction}
Neural Machine Translation (NMT) established on the encoder-decoder framework, where the encoder takes a source sentence as input and encodes it into a fixed-length embedding vector, and the decoder generates the translation sentence according to the encoder embedding, has achieved advanced translation performance in recent years \cite{Kalchbrenner2013recurrent,Sutskever2014Advances,Cho2014Learning,bahdanau2015attention,vaswani2017attention}.
So far, despite the big advance in model architecture, most models keep taking a standard assumption to translate every sentence independently, ignoring the implicit or explicit sentence correlation from document-level contextual clues during translation.

However, document-level information has shown helpful in improving the translation performance from multiple aspects: consistency, disambiguation, and coherence \cite{kuang2018cache}. If translating every sentence is completely independent of document-level context, it will be difficult to keep every sentence translations across the entire document consistent with each other. Moreover, even sentence independent translation may still benefit from document-level clues through effectively disambiguating words by referring to multiple sentence contexts. At last, document-level clues as a kind of global information across the entire text may effectively help generate more coherent translation results compared to the way only adopting local information inside a sentence alone.

There have been few recent attempts to introduce the document-level information into the existing standard NMT models.
Various existing methods \cite{Jean2017context,tiedemann2017context,wang2017exploiting,voita2018context,kuang2018gate,maruf2018document,jiang2019,li-doc2019,Kim2019,Kate2019} focus on modeling the context from the surrounding text in addition to the source sentence.

For the more high-level context, \citet{miculicich2018han} propose a multi-head hierarchical attention machine translation model to capture the word-level and sentence-level information.
The cache-based model raised by \citet{kuang2018cache} uses the dynamic cache and topic cache to capture the inter-sentence connection.
\citet{tan-etal-2019-hierarchical} integrate their proposed Hierarchical Modeling of Global Document Context model (HM-GDC) into the original
Transformer model to improve the document-level translation.

However, most of the existing document-level NMT methods focus on introducing the information of disambiguating global document or the surrounding sentences but fail to comprehend the relationship among the current sentence, the global document information, and the local document information, let alone the refined global document-level clues.

In this way, our proposed model can focus on the most relevant part of the concerned translation from which exactly encodes the related document-level context.

The empirical results indicate that our proposed method significantly improves the BLEU score compared with a strong Transformer baseline and performs better than other related models for document-level machine translation on multiple tasks.

\section{Related Work}
The existing work about NMT on document-level can be divided into two parts: one is how to obtain the document-level information in NMT, and the other is how to integrate the document-level information.

\subsection{Mining Document-level Information}
\citet{tiedemann2017context} propose to simply extend the context during the NMT model training by \emph{concatenation} method.

\citet{wang2017exploiting} obtain all sentence-level representations after processing each sentence by \emph{Document RNN}.
The last hidden state represents the summary of the whole sentence, as well as the summary of the global context is represented by the last hidden state over the sequence of the above sentence-level representations.

\citet{michel2018extreme} propose a simple yet parameter-efficient adaption method that only requires adapting the \emph{Specific Vocabulary Bias} of output softmax to each particular use of the NMT system and allows the model to better reflect distinct linguistic variations through translation.

\citet{Mac2019UsingWD} present a \emph{Word Embedding Average} method to add source context that capture the whole document with accurate boundaries, taking every word into account by an averaging method.

\begin{figure*}[ht]
    \centering
    \includegraphics[width=1.0\textwidth]{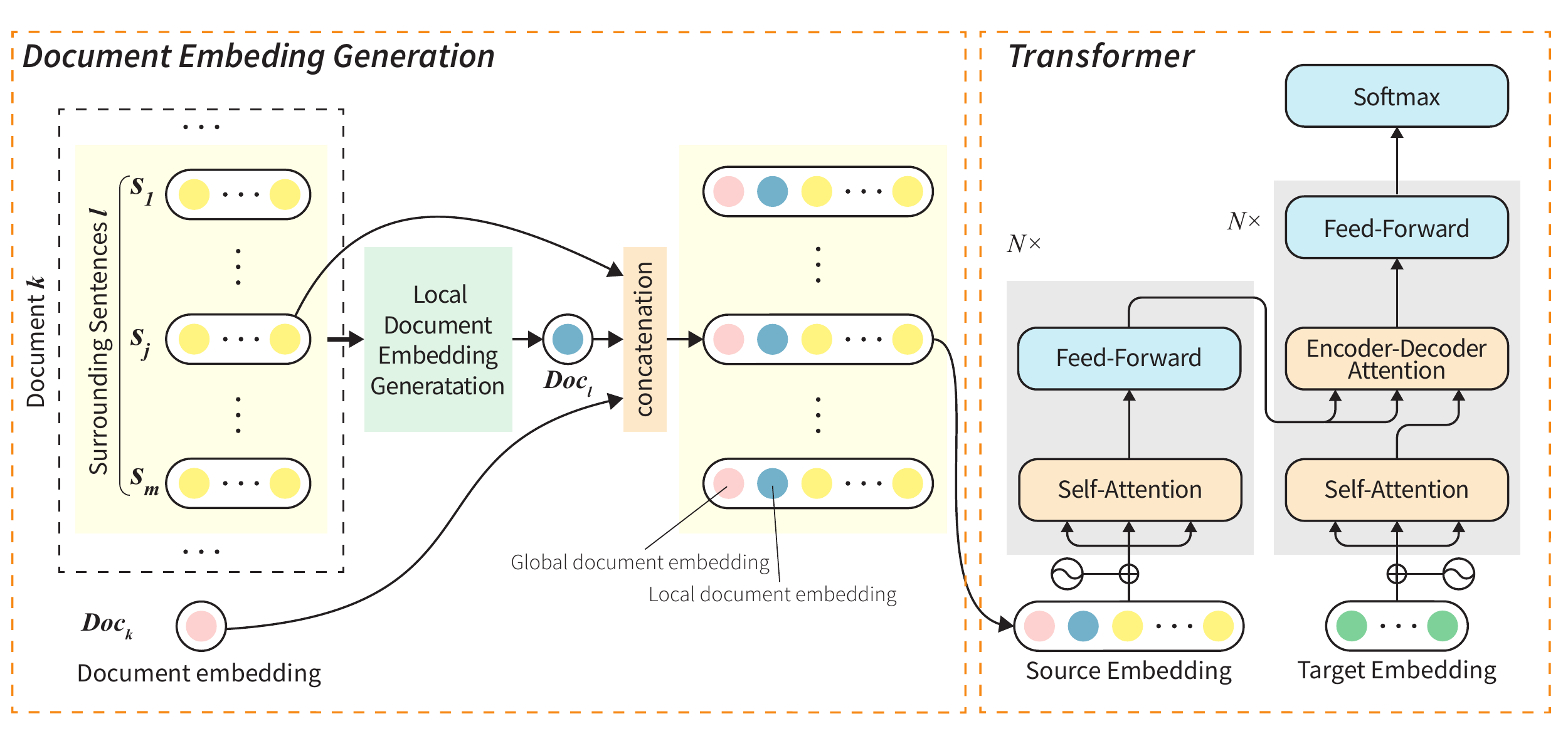}
    \caption{The framework of our model. }
    \label{fig:framework}
\end{figure*}

\subsection{Integrating Document-level Information}

\citet{wang2017exploiting} add the representation of cross-sentence context into the equation of the probability of the next word directly and jointly update the decoding state by the previous predicted word and the source-side context vector.

\citet{tu2017context} introduce a context gate to automatically control the ratios of source and context representations contributions to the generation of target words.\citet{wang2017exploiting} also introduce this mechanism in their work to dynamically control the information flowing from the global text at each decoding step.

\citet{kuang2018gate} propose an inter-sentence gate model, which is based on the attention-based NMT and uses the same encoder to encode two adjacent sentences and controls the amount of information flowing from the preceding sentence to the translation of the current sentence with an inter-sentence gate.

\citet{tu2018learning} propose to augment NMT models by \emph{cache-based neural model} with an external cache to exploit translation history. At each decoding step, the probability distribution over generated words is updated online depending on the translation history retrieved from the cache with a query of the current attention vector.

\citet{voita2018context} introduce the context information into the Transformer \cite{vaswani2017attention} and leave the Transformer's decoder intact while processing the context information on the encoder side.
This \emph{context-aware Transformer model} calculates the gate from the source sentence attention and the context sentence attention, then exploits their gated sum as the encoder output.
\citet{zhang2018improving} also extend the Transformer with a new context encoder to represent document-level context while incorporating it into both the original encoder and decoder by multi-head attention.

\citet{miculicich2018han} propose a \emph{Hierarchical Attention Networks} (HAN) NMT model to capture the context in a structured and dynamic pattern. For each predicted word, it uses word-level and sentence-level abstractions and selectively focuses on different words and sentences.

\citet{tan-etal-2019-hierarchical} propose a global document context model to improve the document-level translation, which is hierarchically extracted from the entire global text with a sentence encoder to model intra-sentence information and a document encoder to model document-level inter-sentence context representation.

\citet{ma-etal-2020-simple} propose a \emph{Flat-Transformer model} with a simple and effective unified encoder that model the bi-directional relationship between the contexts and the source sentences.

\citet{chen-etal-2020-modeling} propose to improve document-level NMT by the means of discourse structure information, and the encoder is based on a HAN \citet{miculicich2018han}. They parse the document to obtain its discourse structure, then introduce a Transformer-based path encoder to embed the discourse structure information of each word and combine the discourse structure information with the word embedding.

Most of the previous works only focus on the context embedding or considering the global text, but our work is able to mine the relationship among input sentences, the whole document, and context sentences like the previous sentences.

\section{Background}
\subsection{Neural Machine Translation}
Given a source sentence $\mathbf{x} = \{x_{1}, ..., x_{i}, ... , x_{S}\}$ in the document to be translated and a target sentence $\mathbf{y}= \{y_{1}, ..., y_{i}, ... , y_{T}\}$, NMT model computes the probability of translation from the source sentence to the target sentence word by word:
\begin{equation}
    P(\mathbf{y}|\mathbf{x}) = \prod^{T}_{i=1}P(y_i|y_{1:i-1},\mathbf{x}),
\end{equation}
where $y_{1:i-1}$ is a substring containing words $y_{1}, ..., y_{i-1}$.
Generally, with the Recurrent Neural Network (RNN), the probability of generating the $i$-th word $y_{i}$ is modeled as:
\begin{equation}
    P(y_i|y_{1:i-1},\mathbf{x}) = \mbox{softmax}(g(y_{i-1},\mathbf{s}_{i-1},\mathbf{c}_i)),
\end{equation}
where $g(\cdot)$ is a nonlinear function that outputs the probability of previously generated word $y_i$, and $\mathbf{c}_i$ is the $i$-th source representation.
Then $i$-th decoding hidden state $\mathbf{s}_i$ is computed as
\begin{equation}
    \mathbf{s}_i = f(\mathbf{s}_{i-1}, y_{i-1}, \mathbf{c}_i).
\end{equation}

For NMT models with an encoder-decoder framework, the encoder maps an input sequence of symbol representations $\mathbf{x}$ to a sequence of continuous representations $\mathbf{z}= \{z_{1}, ..., z_{i}, ... , z_{S}\}$. Then, the decoder generates the corresponding target sequence of symbols $\mathbf{y}$ one element at a time.

\subsection{Transformer Architecture}
Only based on the attention mechanism, \citet{vaswani2017attention} propose a network architecture called Transformer for NMT, which uses stacked self-attention and point-wise, fully connected layers for both encoder and decoder.

The encoder is composed of a stack of $N$ (usually equal to 6) identical layers, and each layer has two sub-layers: (1) multi-head self-attention mechanism, and (2) a simple, position-wise fully connected feed-forward network.

Multi-head attention
in the Transformer allows the model to jointly process information from different representation spaces at distinct positions.
It linearly projects the queries $Q$, keys $K$, and values $V$ $h$ times to $d_k$, $d_k$, and $d_v$ dimensions respectively, then the attention function is performed in parallel, generating $d_v$-dimensional output values, and yielding the final results by concatenating and once again projecting them.
The core of multi-head attention is Scaled Dot-Product Attention and calculated as:
\begin{equation}
    \mbox{Attention}(Q,K,V) = \mbox{softmax}(\frac{QK^T}{\sqrt{d_k}})V.
\end{equation}

The second sub-layer is a feed-forward network, which contains two linear transformations with a ReLU activation in between.

Similar to the encoder, the decoder is also composed of a stack of $N$ identical layers, but it inserts a third sub-layer, which performs multi-head attention over the output of the encoder stack. The Transformer also employs residual connections around each of the sub-layers, followed by layer normalization. Thus, the Transformer is more parallelizable and faster for translating than earlier RNN methods.

\begin{figure*}[ht]
    \centering
    \includegraphics[width=1.0\textwidth]{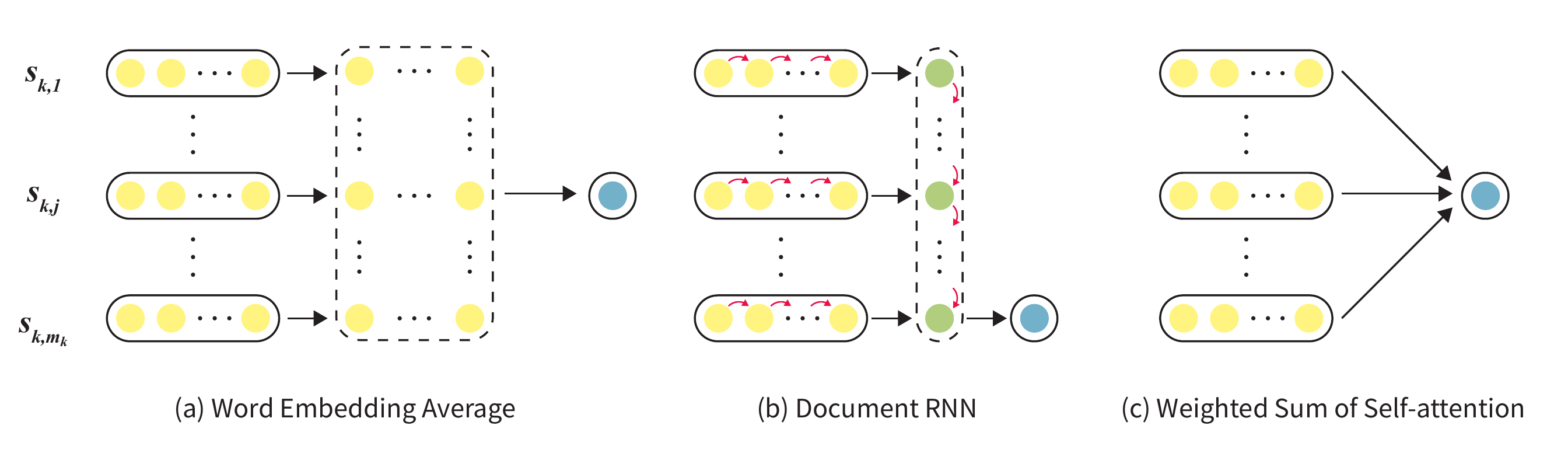}
    \caption{Three methods for calculating the \emph{document embedding}.}
    \label{fig:doc}
\end{figure*}

\begin{center}
\begin{table*}[ht]
\centering
\begin{tabular}{@{}l|rrrr|rrrr|rrrr@{}}
\toprule
        & \multicolumn{4}{c|}{\textbf{Zh-En}}       & \multicolumn{4}{c|}{\textbf{En-Fr}}       & \multicolumn{4}{c}{\textbf{En-De}}        \\
Data & \#Doc & \#Sent & \#Src  & \#Tgt  & \#Doc & \#Sent & \#Src  & \#Tgt  & \#Doc & \#Sent & \#Src  & \#Tgt  \\ \midrule
train   & 1976  & 0.2M & 4.8M   & 5.4M   & 1914  & 0.2M & 5.5M   & 5.9M   & 1705  & 0.2M & 4.9M   & 5.0M   \\ \midrule
dev & 8     & 0.9K   & 23.6K  & 23.4K  & 8     & 0.9K   & 23.6K  & 24.3K  & 8     & 0.9K   & 23.7K  & 24.6K  \\ \midrule
tst10 & 11    & 1.6K   & 35.3K  & 36.0K  & 11    & 1.6K   & 36.3K  & 36.3K  & 11    & 1.6K   & 36.5K  & 37.7K  \\
tst11 & 16    & 1.4K   & 27.0K  & 30.9K  & 16    & 1.4K   & 31.0K  & 33.5K  & 16    & 1.4K   & 30.9K  & 32.7K  \\
tst12 & 15    & 1.7K   & 30.4K  & 34.9K  & 15    & 1.7K   & 35.2K  & 38.7K  & 15    & 1.7K   & 35.1K  & 36.8K  \\
tst13 & 20    & 1.4K   & 31.5K  & 34.1K  & 20    & 1.4K   & 34.3K  & 37.8K  & 16    & 1.0K   & 24.3K  & 24.7K  \\
tst14 & 15    & 1.3K   & 26.5K  & 29.5K  & 15    & 1.3K   & 29.7K  & 33.0K  & 15    & 1.3K   & 29.8K  & 31.0K  \\
tst15 & 12    & 1.2K   & 25.2K  & 27.5K  & 12    & 1.2K   & 27.7K  & 29.6K  & 12    & 1.1K   & 24.1K  & 24.8K  \\
tst-all & 62    & 5.6K   & 0.2M   & 0.2M   & 89    & 8.6K   & 0.2M   & 0.2M   & 85    & 8.1K   & 0.2M   & 0.2M \\ \bottomrule
\end{tabular}
\caption{Detail of training, development and test sets.}
\label{tab:stat}
\end{table*}
\end{center}

\section{Model}
As shown in Figure \ref{fig:framework}, we introduce document-level clues into the NMT training through an embedding method, two types of document embeddings (namely, \emph{global} and \emph{local}) are directly concatenated into the source embedding during training.

\subsection{Document Embedding Generation}
\label{section:doc}
For a document $k$ = $\{ \mathbf{x}_1,...,\mathbf{x}_j,...,\mathbf{x}_{m_k}\} $, the sentence in this document is represented $\mathbf{x}_j =\{ x_1,...,x_i,...,x_{n_j}\}$, the word vector in the sentence is denoted as $x_i$.

The \emph{global} document embedding is obviously generated from the whole document $k$ in the corpus with fine-defined document boundaries.
While generating the \emph{local} document embedding, we consider the surrounding sentences $\{\mathbf{x}_j\}^{m}_{j=1}$ of the current sentence $\mathbf{x}_j$ as the document $l$.

In this paper, we consider the following methods to obtain the document embedding.

\subsubsection{Word Embedding Average}
According to \citet{Mac2019UsingWD}, we consider the \emph{document embedding} of a document $k$ by averaging all $N$ word vectors $x$ in this document and therefore has the same dimension.
\begin{equation}
    \text{Doc}_k = \frac{1}{N} \sum_{i=1}^{N}x_{i,k}
\end{equation}

\subsubsection{Document RNN}
Inspired by \citet{wang2017exploiting} which proposes a cross-sentence context-aware RNN approach to produce the context representation.
Firstly, the sentence RNN reads the corresponding words $x_{i,j}$ in the sentence $\mathbf{x}_j$ sequentially and updates its hidden state by
\begin{equation}
    h_{i,j} = f(h_{i-1,j}, x_{i,j})
\end{equation}
where $f(\cdot)$ is an activation function, and $h_{i,j}$ is the hidden state at time $i$. The last state $h_{n_j,j}$ represents the summary of the whole sentence $\mathbf{x}_j$ and denoted as $\text{Sent}_j$, i.e. $h_{n_j,j} \equiv \text{Sent}_j$.

Then, all sentence-level representations are fed into document RNN as follows:
\begin{equation}
    H_{j,k} = f(H_{j-1,k}, \text{Sent}_{j,k})
\end{equation}
where $h_{j,k}$ is the recurrent state at time $j$ in the document $k$. Similarly, we use the last hidden state to indicate the summary of the global document, i.e. $H_{j,k} \equiv \text{Doc}_k$.

\subsubsection{Weighted Sum of Self-attention}
The input sentence $\mathbf{x}_j$ first goes through a multi-head attention layer to encode the contextualized information to each word representation:
\begin{equation}
    \mathbf{X}^{(n)}_j = \mbox{MultiHead}(\mathbf{X}^{(n-1)}_j,\mathbf{X}^{(n-1)}_j,\mathbf{X}^{(n-1)}_j) ,
\end{equation}
where $\mathbf{X}^{(0)}_j = \mathbf{x}_j$.
Each word representation is as a vector $s \in \mathbb{R}^d $, where $d$ is the size of hidden state in $\mbox{MultiHead}$ function.

Then we compute the compatibility function using a feed-forward network FFN with a single hidden layer:
\begin{equation}
\mathbf{X}^{(n)}_j = [\mbox{FNN}(\mathbf{X}^{(n)}_{1,j});...;\mbox{FNN}(\mathbf{X}^{(n)}_{n,j})]
\end{equation}
where $\mathbf{X}^{(n)}_j\in \mathbb{R}^{i \times d} $ is the representation of the source sentence $\mathbf{x}_j$ at the $n$-th layer $(n = 1,..,N)$.
We treat the output at the last layer as the representation of the input sentence $\mathbf{x}_j$ i.e. $\mathbf{X}^{(N)}_j \equiv \text{Sent}_j$.

Finally, we calculated the \emph{document embedding} by acquiring the weighted sum of all sentence embeddings in the document $k$:
\begin{equation}
\text{Doc}_k = \sum_{j=1}^{m_k}\alpha_j \text{Sent}_{j,k}.
\end{equation}
where $\alpha_j$ is the weight of each sentence embedding $\text{Sent}_{j,k}$ trained by the model.

\begin{table*}[ht]
\centering
\begin{tabular}{@{}ll|r|rrrrrr@{}}
\toprule
\textbf{Language} & \textbf{Model}        & \textbf{tst-all} & \textbf{tst10} & \textbf{tst11} & \textbf{tst12} & \textbf{tst13} & \textbf{tst14} & \textbf{tst15} \\ \midrule
                  & RNN Search*           & 16.80            & 12.96            & 17.40            & 15.39            & 15.89            & 13.15            & 15.85            \\
                  & Baseline              & 19.50            & 16.44            & 21.24            & 18.93            & 20.05            & 17.99            & 21.04            \\
Zh-En             & Enhanced              & 19.98            & 16.80            & 21.56            & 19.31            & 20.67            & \textbf{18.41}   & 21.75            \\
                  & Enhanced+global       & 20.13            & 16.63            & 21.55            & 19.16            & 20.88            & 18.02            & 21.79            \\
                  & Enhanced+global+local & \textbf{20.18}   & \textbf{16.84}   & \textbf{21.68}   & \textbf{19.36}   & \textbf{20.90}   & 18.38            & \textbf{21.89}   \\\midrule
                  & RNN Search*           & 37.13            & 31.76            & 38.18            & 36.91            & 35.59            & 33.28            & 32.38            \\
                  & Baseline              & 39.92            & 36.03            & 43.07            & 42.83            & 40.34            & 38.10            & 38.70            \\
En-Fr             & Enhanced              & 39.98            & 36.47            & 43.51            & 43.07            & 41.23            & 38.35            & 38.85            \\
                  & Enhanced+global       & 41.25            & \textbf{37.57}   & 43.97            & 44.49            & \textbf{42.34}   & 39.28            & 39.17            \\
                  & Enhanced+global+local & \textbf{41.46}   & 37.30            & \textbf{44.48}   & \textbf{44.71}   & 42.25            & \textbf{39.44}   & \textbf{39.35}   \\\midrule
                  & RNN Search*           & 25.28            & 23.33            & 25.46            & 22.18            & 24.68            & 20.56            & 22.59            \\
                  & Baseline              & 27.94            & 28.12            & 30.41            & 26.67            & 29.11            & 25.28            & 27.24            \\
En-De             & Enhanced              & 28.46            & 28.39            & 30.05            & 28.11            & 30.33            & 26.39            & 28.59            \\
                  & Enhanced+global       & 29.03            & \textbf{29.56}   & \textbf{31.21}   & 27.93            & 30.72            & 26.21            & 28.33            \\
                  & Enhanced+global+local & \textbf{29.21}   & 29.44            & 31.09            & \textbf{28.48}   & \textbf{30.99}   & \textbf{26.76}   & \textbf{28.92}   \\ \bottomrule
\end{tabular}
\caption{Performance (BLEU scores) comparison on the different datasets. ``Enhanced'' indicates the \emph{Enhanced} model defined in Section \ref{section:doc_inte} benefiting from the word embeddings extracted from the \emph{Baseline} Model. The \emph{global} and \emph{local} embeddings are generated by ``Word Embedding Average'' method and ``Document RNN'' method respectively. The proposed methods were significantly better than the baseline Transformer at significance level $p$-value$<$0.05~. The scores in bold indicate the best ones on the same dataset.}
\label{tab:result}
\end{table*}

\begin{table*}[ht]
\centering

\begin{tabular}{@{}l|lll@{}}
\toprule
                                                     & Zh-En     & Zh-En     & En-De     \\
Corpus                                               & IWSLT2015 & IWSLT2017 & IWSLT2017 \\ \midrule
HAN \citep{miculicich2018han}                        & 17.79     & -         & -         \\
HM-GDC \citep{tan-etal-2019-hierarchical}            & -         & 17.63     & -         \\
Flat-Transformer \citep{ma-etal-2020-simple}         & -         & -         & 26.61     \\
Transformer+HAN+DS \citep{chen-etal-2020-modeling}   & -         & -         & 24.84     \\
Our model(\emph{Enhanced}+global+local (avg+rnn))    & \textbf{19.40}     & \textbf{20.61}     & \textbf{29.21}     \\ \bottomrule
\end{tabular}

\caption{Comparison with the related works.}
\label{tab:comparsion}
\end{table*}

\subsection{Document Embedding Integration}
\label{section:doc_inte}

At first, we train a baseline Transformer model (noted \emph{Baseline} model) on a standard corpus that does not contain any document-level information and extract word embeddings from it.
Then, we train an enhanced model (noted \emph{Enhanced} model) benefiting from these extracted word embeddings.
This process can be treated as \emph{pre-training} of word embeddings. The \emph{Enhanced} model directly adopts the pre-trained embeddings from the \emph{Baseline} model, namely, word embeddings will be fixed during its model training.

In our model, we calculated the \emph{global} document embedding $\text{Doc}_k$ for each document $k$ using \emph{Word Embedding Average} method.
It should be noted that the enhanced model does not fine-tune its embeddings to preserve the relationship between words and document vectors during training.

Because of the limitation of GPU memory, it is impractical to feed all the word vectors in the document $k$ into the Model and calculate the \emph{global} document embedding using \emph{Document RNN} and \emph{Weighted Sum of Self-attention} method, which needs train the hidden variables to generate the document embedding.
On the basis of this restriction, we generate the \emph{local} document embedding from the \emph{surrounding sentences} $l=\{\mathbf{x}_j\}^{m}_{j=1}$ during training and denote it as $\text{Doc}_l$.
In practice, we using the input sentences in a mini-batch as the \emph{surrounding sentences}.
Thus,the source embedding $\mathbf{s}$ for input sentence $\mathbf{x}$ can be represented as follows:
\begin{equation}
    \mathbf{s} = \{\text{Doc}_k, \text{Doc}_l, x_1, x_i,...,x_n\}
\end{equation}

In the case that it needs to ensemble $N$ document embeddings, for example to ensemble ``Document RNN'' and ``Weighted Sum of Self-attention'', we calculate the weighed sum of them:
\begin{equation}
\text{Doc} = \sum_{i=1}^{N}\beta_i \text{Doc}_{i}
\end{equation}
where $\beta_i$ is the weight of each document embedding $Doc_i$ trained by the model.

\begin{table*}[ht]
\centering
\begin{tabular}{@{}lll|r|rrrrrr@{}}
\toprule
\textbf{Enhanced+} & \textbf{Global} & \textbf{Local} & \textbf{tst-all} & \textbf{tst10} & \textbf{tst11} & \textbf{tst12} & \textbf{tst13} & \textbf{tst14} & \textbf{tst15} \\ \midrule
                   & /               & /              & 19.98            & 16.80          & 21.56          & 19.31          & 20.67          & \textbf{18.41} & 21.75          \\
                   & avg             & /              & 20.13            & 16.63          & 21.55          & 19.16          & 20.88          & 18.02          & 21.79          \\
                   & /               & avg            & 19.96            & 16.91          & 21.50          & 19.08          & 20.89          & 18.12          & 21.75          \\
                   & /               & rnn            & 19.87            & 16.69          & 21.65          & 19.36          & 20.68          & 18.23          & 21.62          \\
Zh-En              & /               & attn           & 20.09            & 16.70          & 21.54          & 19.22          & 20.76          & 18.20          & 21.76          \\
                   & avg             & avg            & 20.12            & 16.73          & 21.57          & 19.25          & 20.79          & 18.23          & 21.78          \\
                   & avg             & rnn            & 20.18            & 16.84          & 21.68          & 19.36          & \textbf{20.90} & 18.34          & \textbf{21.89} \\
                   & avg             & attn           & 20.07            & 16.68          & 21.52          & 19.20          & 20.74          & 18.18          & 21.74          \\
                   & avg             & rnn+attn       & \textbf{20.23}   & \textbf{16.93} & \textbf{21.70} & \textbf{19.42} & 20.86          & 18.04          & 21.76          \\ \midrule
                   & /               & /              & 39.98            & 36.47          & 43.51          & 43.07          & 41.23          & 38.35          & 38.85          \\
                   & avg             & /              & 41.25            & \textbf{37.57} & 43.97          & 44.49          & 42.34          & 39.28          & 39.17          \\
                   & /               & avg            & 41.32            & 37.18          & 44.13          & 44.63          & \textbf{42.39} & 39.19          & 39.22          \\
                   & /               & rnn            & 40.68            & 37.09          & 43.81          & 43.85          & 41.85          & 38.88          & 39.08          \\
En-Fr              & /               & attn           & 40.58            & 36.98          & 43.70          & 43.74          & 41.75          & 38.78          & 38.97          \\
                   & avg             & avg            & 41.01            & 36.85          & 44.13          & 44.42          & 41.83          & 38.59          & 38.74          \\
                   & avg             & rnn            & \textbf{41.46}   & 37.30          & \textbf{44.48} & \textbf{44.71} & 42.25          & \textbf{39.44} & \textbf{39.35} \\
                   & avg             & attn           & 40.89            & 37.36          & 43.84          & 43.77          & 41.31          & 39.00          & 39.08          \\
                   & avg             & rnn+attn       & 40.95            & 37.35          & 44.07          & 44.11          & 42.12          & 39.15          & 39.34          \\ \midrule
                   & /               & /              & 28.46            & 28.39          & 30.05          & 28.11          & 30.33          & 26.39          & 28.59          \\
                   & avg             & /              & 29.03            & 29.56          & 31.21          & 27.93          & 30.72          & 26.21          & 28.33          \\
                   & /               & avg            & 28.92            & 28.88          & 30.46          & 27.47          & \textbf{31.75} & 26.48          & 28.69          \\
                   & /               & rnn            & 29.18            & \textbf{29.77} & \textbf{31.42} & 28.11          & 30.66          & 26.20          & 27.86          \\
En-De              & /               & attn           & 28.46            & 28.69          & 30.35          & 27.74          & 30.24          & 26.02          & 28.18          \\
                   & avg             & avg            & 28.91            & 29.14          & 30.79          & 28.18          & 30.69          & 26.46          & 28.62          \\
                   & avg             & rnn            & \textbf{29.21}   & 29.44          & 31.09          & \textbf{28.48} & 30.99          & \textbf{26.76} & \textbf{28.92} \\
                   & avg             & attn           & 28.47            & 28.70          & 30.36          & 27.75          & 30.25          & 26.03          & 28.19          \\
                   & avg             & rnn+attn       & 28.73            & 28.96          & 30.61          & 28.00          & 30.51          & 26.28          & 28.44          \\ \bottomrule
\end{tabular}
\caption{Ablation study on the \emph{Enhanced} model with different document embeddings. The tag \emph{avg}, \emph{rnn} and \emph{attn} mean that the document embedding is generated by ``Word Embedding Average'' method, ``Document RNN'' and ``Weighted Sum of Self-attention'' method.}
\label{tab:ablation}
\end{table*}

\begin{table*}[ht]
\centering
\begin{tabular}{@{}ll@{}}
\toprule
\multirow{4}{*}{Context sentences} & \emph{Zhídào \textcolor{red}{shànggè shìjì 80 niándài}, zhège   yúchǎng shì āgēntíng rén guǎnxiá de}. \\
                                   & \emph{(until \textcolor{red}{the 1980s}, the farm was in the hands of the Argentinians.)}              \\
                                   & \emph{Tāmen zài zhèlǐ yǎng niú nàgè shíhòu zhèlǐ jīběn shàng shì shīdì.}              \\
                                   & \emph{(they raised beef cattle on what was essentially wetlands.)}                    \\ \midrule
Source sentence                    & \emph{Dāngshí tāmen bǎ shuǐ chōu zǒu}.                                                \\ \midrule
Reference sentence                 & \emph{they \textcolor{blue}{did} it by draining the land.}                                             \\
Transformer model                  & \emph{and then they take the water off.}                                             \\
\textbf{Our model}                 & \emph{and then they \textcolor{blue}{pulled} the water out.}                                           \\ \bottomrule
\end{tabular}
\caption{The first example of the translation result. The context sentences are the previous 2 sentences before the source sentence and words in red from context indicate the heuristic clues for better translation. The Chinese sentences are converted to Pinyin version and the English translation have been provided.}
\label{tab:example1}
\end{table*}

\begin{table*}[ht]
\centering
\begin{tabular}{@{}ll@{}}
\toprule
\multirow{4}{*}{Context sentences} & \emph{Duì wǔzhǒng fāngfǎ zhōng de měi yīzhǒng, wǒmen dōu xūyào zhìshǎo 100 rén de tuánduì.}       \\
                                   & \emph{(in each of these five paths, we need at least a hundred people.)}                        \\
                                   & \emph{Lǐmiàn de hěnduō rén, nǐ huì juédé tāmen hěn fēngkuáng, zhè jiù duìle.}                     \\
                                   & \emph{(and a lot of them, you'll look at and say, "They're crazy ." that's good.)}            \\ \midrule
Source sentence                    & \emph{Wǒ rènwéi, zài TED tuánduì lǐ yǐjīng yǒu hěnduō rén kāishǐ zhìlì yú cǐ.}                    \\ \midrule
Reference sentence                 & \emph{\textcolor{blue}{and}, I think, here in the TED group, we have many people who are already pursuing this.} \\
Transformer model                  & \emph{I think there are so many people in the TED community that are working on this.}           \\
\textbf{Our model}                 & \emph{\textcolor{blue}{and} I think there‘s a lot of people in the TED community who have been working on this.}   \\ \bottomrule
\end{tabular}
\caption{The second example of the translation result. The context sentences are the previous 2 sentences before the source sentence. The Chinese sentences are converted to Pinyin version and the English translation have been provided.}
\label{tab:example2}
\end{table*}


\section{Experiments}
\subsection{Setup}
\paragraph{Data}
As we focus on document-level NMT, it poses a document-annotation requirement on the evaluation dataset that needs for well defined document boundaries marking each sentence with its global document tag.
Thus we train and evaluate our model on the corpus from the TED Talks on three language pairs, i.e., Chinese-to-English (Zh-En), English-to-French(En-Fr), and English-to-German(En-De).
The TED talk documents are the parts of the IWSLT2017 Evaluation Campaign Machine Translation task\footnote{\url{https://wit3.fbk.eu/mt.php?release=2017-01-trnted}}.
We use \emph{dev2010} as the development set and combine the \emph{tst2010-2015} as the test set. The statistics of the corpora are listed in Table \ref{tab:stat}.

\paragraph{Data preprocessing}
The English and Spanish datasets are tokenized by \emph{tokenizer.perl} and truecased by \emph{truecase.perl} provided by MOSES\footnote{\url{https://github.com/moses-smt/mosesdecoder}}, a statistical machine translation system proposed by \citet{Koehn2007moses}. The Chinese corpus is tokenized by \emph{Jieba} Chinese text segmentation\footnote{\url{https://github.com/fxsjy/jieba}}. Words in sentences are segmented into subwords by Byte-Pair Encoding (BPE) \citet{sennrich2016bpe} with 32k BPE operations.

\paragraph{Model Configuration}
We use the Transformer proposed by \citet{vaswani2017attention} as our baseline and implement our work using the THUMT, an open-source toolkit for NMT developed by the Natural Language Processing Group at Tsinghua University \cite{zhang2017THUMT}\footnote{\url{https://github.com/thumt/THUMT}}.
We follow the configuration of the Transformer ``base model'' described in the original paper \cite{vaswani2017attention}.
Both encoder and decoder consist of 6 hidden layers each. All hidden states have 512 dimensions, 8 heads for multi-head attention, and the training batch contains about 6,520 source tokens.
We use the original regularization and optimizer in Transformer \cite{vaswani2017attention}.
Finally, we evaluate the performance of the model by BLEU score \cite{papineni2002bleu} using \emph{multi-bleu.perl} on the \emph{tokenized} text.

\subsection{Translation Performance}
Table \ref{tab:result} demonstrates the BLEU scores for different models on multiple corpora.
The \emph{RNN} model is a re-implemented attention-based NMT system RNNSearch* \cite{hinton2012rnn} and Transformer \cite{vaswani2017attention} using THUMT kit. The \emph{Baseline} Model is also the \emph{pre-trained} model mentioned in the Section \ref{section:doc}.

The results in Table \ref{tab:result} demonstrate that our model is significantly better than the baseline Transformer at significance level $p$-value$<$0.05.
The \emph{global} embedding (generated by the ``Word Embedding Average'' method) and the \emph{local} embedding (generated by ``Document RNN'' method) in our model can effectively exploit the document-level information from the global text and the surrounding sentences and improve the performance of the \emph{Enhanced} model.

The \emph{Enhanced} model trained with the embedding form the \emph{Baseline} model outperforms the \emph{Baseline} model by 0.48 BLEU point on the Zh-En dataset, 0.06 BLEU point on the En-Fr dataset, and 0.52 BLEU point on the En-De dataset.
When we add the \emph{global} document embedding to the \emph{Enhanced} model, the \emph{Enhanced+global} model, surpassed the \emph{Baseline} model by 0.73 BLEU points and Zh-En, 1.40 BLEU points on En-Fr and 1.24 BLEU points on En-De.
Moreover, after taking the \emph{local} document embedding into account, the \emph{Enhanced+global+local} model achieves the gains of 1.11 BLEU point, 1.54 BLEU point, and 1.27 BLEU point on these three datasets individually over the \emph{Baseline} model.

\subsection{Comparison with the related work}
We also compared our proposed method on the corpus mentioned in the Hierarchical Attention Networks (HAN) NMT \cite{miculicich2018han} model, Hierarchical Modeling of Global Document Context methods (HM-GDC) \cite{tan-etal-2019-hierarchical}, Flat-Transformer \cite{ma-etal-2020-simple}, and document-level NMT based on a HAN with discourse structure information (Transformer+HAN+DS) model\citet{chen-etal-2020-modeling} and the results in Table \ref{tab:comparsion} show that our model significantly outperforms the related work.

\subsection{Ablation Study}
We investigate the impact of different document embedding methods by removing one or more of them.
As shown in Table \ref{tab:ablation}, all of the components greatly contribute to the performance of our proposed model.
If we remove any document embedding in the \emph{Enhanced} model, the performance drops dramatically.
Such results indicate that both the \emph{global} document embedding and the \emph{local} document embedding play an important and complementary role in our model.
For the \emph{local} document embedding, we compare different document embedding generation methods and find out that the ``Document RNN'' method has great effects on \emph{Enhanced} model.

\subsection{Translation Quality}
We also provide examples to illustrate what do these document embeddings capture.
Table \ref{tab:example1} shows the first example, which is extracted from line 111 of TED Talks (Zh-En). The source sentence does not involve any information to indicate the time status, but the context sentences mention the time information ``\emph{shànggè shìjì 80 niándài}'' (which means ``the 1980s'' in English). Thus our model can recognize the past tense of the source sentence exactly.
Table \ref{tab:example2} demonstrates the translation example form line 549 of TED Talks (Zh-En) as the second example. Although the source sentence does not contain any word to represent the discourse relationship with previous contexts, our model is able to infer discourse relationship and add the connection word to make the translation more fluent.

\section{Conclusion} 

In this paper, we explore more comprehensive document-level neural machine translation.
Assuming that document-level clues are indeed helpful for better translation, it is kept an open problem for finding a good way to effectively introduce such helpful clues into sentence-independent NMT. Taking document embedding as our default representation for document-level clues, we distinguish two types of document embeddings, the global and the local, which targetedly capture both the general information in the whole document scope and the specific detailed information in the surrounding text. For the concerned document-level NMT, we for the first time survey multiple ways for generating document embeddings and conduct extensive experiments. Taking a strong Transformer baseline, our experimental results show that our global and local document embeddings may effectively enhance the baseline systems, showing that more sufficient and richer document clues indeed greatly help standard sentence-independent NMT.



\bibliography{aaai21}

\end{document}